\documentclass[10pt,twocolumn,letterpaper]{article}

\usepackage{iccv}
\usepackage{times}
\usepackage{epsfig}
\usepackage{graphicx}
\usepackage{amsmath}
\usepackage{amssymb}
\usepackage{algorithm}
\usepackage[noend]{algpseudocode}
\usepackage{siunitx}
\usepackage{multirow}
\usepackage{interval}
\usepackage{adjustbox}
\usepackage{caption}
\usepackage{subcaption}
\usepackage{textcomp}

\usepackage[pagebackref=true,breaklinks=true,letterpaper=true,colorlinks,bookmarks=false]{hyperref}

\def\iccvPaperID{308} 

\iccvfinalcopy
\begin{document}

\title{Parallel Structure from Motion from Local Increment to Global Averaging}

\author{Siyu Zhu \and Tianwei Shen \and Lei Zhou \and Runze Zhang \and Jinglu Wang \and Tian Fang \and Long Quan\\
The Hong Kong University of Science and Technology\\
{\tt\small \{szhu,tshenaa,lzhouai,rzhangaj,jwangae,tianft,quan\}@cse.ust.hk}
}

\maketitle

\begin{figure}
\vspace{-2.1in}
\begin{minipage}[b]{1\textwidth}
\includegraphics[width=1.0\linewidth]{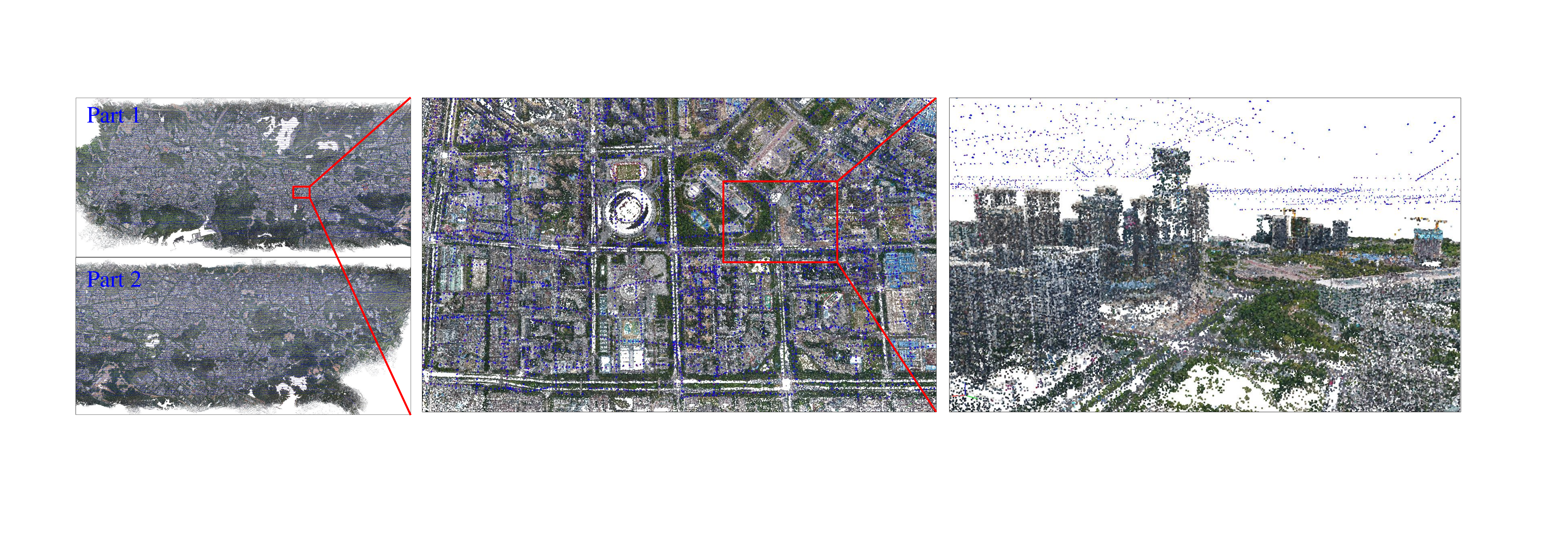}
\caption{Our scalable and parallel SfM system recovers accurate and consistent 1.21 million camera poses (marked by blue dots) and 1.68 billion sparse 3D points of a typical medium-sized city from 50 megapixel high-resolution images.
The figures from left to right zoom successively closer to the final representative buildings.
}
\label{fig:teaser}
\end{minipage}
\vspace{-5.4mm}
\end{figure}

\begin{abstract}
\vspace{-2.3mm}
In this paper, we tackle the accurate and consistent Structure from Motion (SfM) problem, in particular camera registration, far exceeding the memory of a single computer in parallel.
Different from the previous methods which drastically simplify the parameters of SfM and sacrifice the accuracy of the final reconstruction,
we try to preserve the connectivities among cameras by proposing a camera clustering algorithm to divide a large SfM problem into smaller sub-problems in terms of camera clusters with overlapping.
We then exploit a hybrid formulation that applies the relative poses from local incremental SfM into a global motion averaging framework and produce accurate and consistent global camera poses.
Our scalable formulation in terms of camera clusters is highly applicable to the whole SfM pipeline including track generation, local SfM, 3D point triangulation and bundle adjustment.
We are even able to reconstruct the camera poses of a city-scale data-set containing more than one million high-resolution images
with superior accuracy and robustness evaluated on benchmark, Internet, and sequential data-sets.
\vspace{-4.4mm}
\end{abstract}


\begin{figure*}[th!]
\centering
\includegraphics[width=1.0\linewidth]{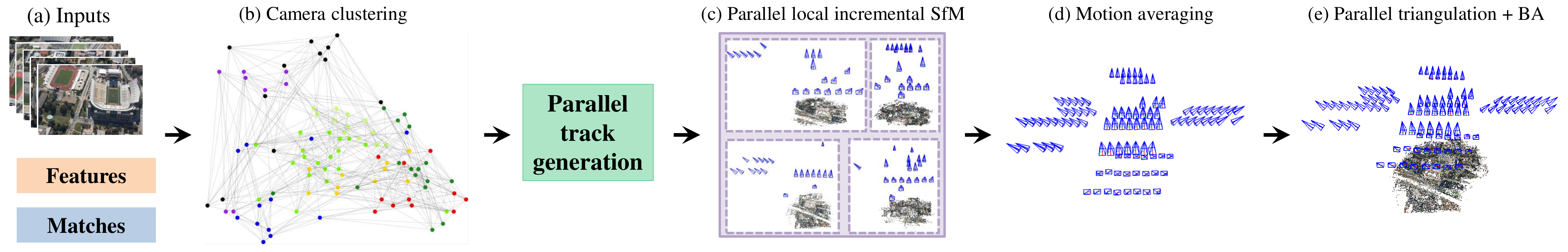}
\caption{The pipeline of our system. The figures from left to right are respectively (a) the input images, features and matches, (b) the camera clusters after the camera clustering algorithm (section~\ref{sec:camera_clustering}), (c) the structure and motion of different camera clusters recovered by parallel local incremental SfM (section~\ref{sec:local_sfm}), (d) the averaged global camera poses after motion averaging (section~\ref{sec:motion_averaging}), and (e) the camera poses and sparse 3D points after parallel 3D point triangulation and parallel bundle adjustment (section~\ref{sec:bundle_adjustment}).}
\label{fig:pipeline}
\vspace{-6mm}
\end{figure*}

\section{Introduction}
The prestigious large-scale SfM methods~\cite{agarwal2011,frahm2010,heinly2015,klingner2013,schonberger2015,snavely2008} have already provided ingenious designs in feature extraction~\cite{lowe2004,siftgpu07wu}, 
overlapping image detection~\cite{agarwal2011,frahm2010,heinly2015,nister2006scalable}, feature matching and verification~\cite{wu2013}, 
and bundle adjustment~\cite{eriksson2016,ni2007,wu2011}.
However, the large-scale accurate and consistent camera registration problem has not been completely solved, not to mention in a parallel fashion.

To fit a whole camera registration problem into a single computer, previous works~\cite{agarwal2011,frahm2010,heinly2015,schonberger2015,snavely2008} generally drastically discard the connectivities among cameras and tracks by first building a skeletal geometry of iconic images~\cite{li2008} and registering the remaining cameras with respect to the skeletal reconstruction.
The other approaches~\cite{havlena2010,moulon2013,resch2015,sweeney2017,toldo2015} generate exclusive camera clusters for partial reconstruction and finally merge them together.
Such losses of camera-to-camera connectivities remarkably decrease the accuracy and consistency of the final reconstruction.
Instead, this work tries to preserve the camera-to-camera connectivities and their corresponding tracks for a highly accurate and consistent reconstruction.
We propose an iterative camera clustering algorithm that splits the original SfM problem into several smaller sub-problems in terms of clusters of cameras with overlapping.
We then exploit this scalable framework to solve the whole SfM problem,
including track generation, local SfM, 3D point triangulation and bundle adjustment far exceeding the memory of a single computer in a parallel scheme.

To obtain the global camera poses from partial sparse reconstructions,
the hybrid SfM methods~\cite{bhowmick2014,sweeney2017} directly use similarity transformations to roughly merge clusters of cameras together and possibly lead to inconsistent camera poses across clusters.
Others~\cite{farenzena2009,havlena2010,lhuillier2005,resch2015,toldo2015} hierarchically merge camera pairs and triplets and are sensitive to the order of the merging process.
Given that the camera-to-camera connectivities are preserved by our clustering algorithm at all possible,
we instead apply the accurate and robust relative poses from incremental SfM~\cite{agarwal2011,pollefeys2004,schoenberger2016sfm,snavely2006,wu2013} 
to the global motion averaging framework~\cite{arie-nachimson2012,brand2004,carlone2015,chatterjee2013,cuibmvc2015,cui2015,goldstein2016,govindu2001,govindu2004,hartley2013,martinec2007,ozyesil2015,sinha2010},
and obtain the global camera poses.

The contributions of our approach are three-fold.
First, we introduce a highly scalable framework to handle SfM problems exceeding the memory of a single computer. 
Second, a camera clustering algorithm is proposed to guarantee that sufficient camera-to-camera connectivities and corresponding tracks are preserved in camera registration.
Finally, we present a hybrid SfM method that uses relative motions from incremental SfM to globally average the camera poses and achieve the state-of-the-art accuracy evaluated on benchmark data-sets~\cite{strecha2008benchmarking}.
To the best of my knowledge, ours is the first pipeline able to reconstruct highly accurate and consistent camera poses from more than one million high-resolution images in a parallel manner.

\section{Related Works}
Based on an initial camera pair, 
the well-known incremental SfM method~\cite{snavely2006} and its derivations~\cite{agarwal2011,pollefeys2004,schoenberger2016sfm,wu2013} progressively recover the pose of the ``next-best-view" by carrying out perspective-three-point (P3P)~\cite{kneip2011} combined with RANSAC~\cite{fischler1981} and non-linear bundle adjustment~\cite{triggs1999} to effectively remove outlier epipolar geometry and feature correspondences.
However, frequent intermediate bundle adjustment leads to incredible time consumption and drifted optimization convergence, especially on large-scale data-sets.
In contrast, the global SfM methods~\cite{arie-nachimson2012,brand2004,carlone2015,chatterjee2013,cuibmvc2015,cui2015,goldstein2016,govindu2001,govindu2004,hartley2013,martinec2007,ozyesil2015,sinha2010} solve all the camera poses simultaneously from the available relative poses,
the computation of which is highly parallel,
and can effectively avoid drifting errors.
Compared with incremental SfM methods,
global SfM methods are, however, more sensitive to possible erroneous epipolar geometry despite the various delicate designs of epipolar geometry filters~\cite{cui2015,govindu2006,heinly2014,jiang2013,moulon2013,roberts2011,wilson2013,wilson2014,zach2008,zach2010}.

In this paper, we embrace the advantages of both incremental and global SfM methods and exploit a hybrid SfM formulation. 
The previous hybrid methods~\cite{farenzena2009,havlena2010,lhuillier2005,resch2015,toldo2015} are limited to small-scale or sequential data-sets.
Havlena \etal~\cite{havlena2010} form the final 3D model by merging atomic 3D models from camera triples together, 
while the merging process is not robust depending solely on common 3D points.
Bhowmick\etal~\cite{bhowmick2014} directly estimate the similarity transformations to combine camera clusters but produce possibly inconsistent camera poses across clusters.
The work in~\cite{sweeney2017} incrementally merges multiple cameras while suffering from severe drifting errors.
In contrast, we apply the robust relative poses from partial reconstruction by local incremental SfM to the global motion averaging framework and provide highly consistent and accurate camera poses.
The work in~\cite{sweeney2017} optimizes the relative poses by solving a single global optimization problem rather than multiple local problems, and suffers from scalability in very large-scale data-sets.

To tackle the scalability problem of large-scale SfM,
previous works generally exploit a skeletal~\cite{snavely2008} or simplified graph~\cite{agarwal2011,frahm2010,heinly2015,schonberger2015}  of iconic images~\cite{li2008}.
Although millions of densely sampled Internet images can be roughly registered, 
numerous geometry connectivities are discarded.
Therefore, such approaches can hardly guarantee a highly accurate and consistent reconstruction in our scenario consisted of uniformly captured high-resolution images.
The hybrid SfM pipelines~\cite{bhowmick2014,havlena2010} employing exclusive clusters of cameras lose a large number of connectivities among cameras and tracks during the cluster partition as well.
Instead, our proposed camera clustering algorithm produces clusters of cameras with overlapping guaranteeing that sufficient camera-to-camera connectivities and corresponding tracks are validated and preserved in camera registration and consequently achieve superior reconstruction accuracy and consistency.


\vspace{-1mm}
\section{Scalable Formulation}
\subsection{Preliminary}
We start with a given set of images $\mathcal{I}=\{I_i\}$, their corresponding SIFT~\cite{lowe2004} features $\mathcal{F} = \{\mathcal{F}_i\}$ and matching correspondences $\mathcal{M}=\{\mathcal{M}_{ij}\,|\,\mathcal{M}_{ij}\subset\mathcal{F}_i\times\mathcal{F}_j,\, i \neq j\}$ where $\mathcal{M}_{ij}$ is a set of inlier feature correspondences verified by epipolar geometry~\cite{hartley2003multiple} between two images $I_i$ and $I_j$.
Each image $I_i$ is associated with a camera $C_i\in\mathcal{C}$.
The target of this paper is then to compute the global camera poses of all the cameras $\mathcal{C}=\{C_i\}$ with projection matrices denoted by $\{\mathbf{P}_i|\mathbf{P}_i=\mathbf{K}_i[\mathbf{R}_i|-\mathbf{R}_i\mathbf{c}_i]\}$.

\subsection{Camera Clustering}\label{sec:camera_clustering}
As the problem of SfM, in particular camera registration, scales up, the following two problems emerge.
First, the problem size gradually exceeds the memory of a single computer.
Second, the high degree parallelism of our distributed computing system can hardly be fully utilized.
We therefore introduce a camera clustering algorithm to split the original SfM problem into several smaller manageable sub-problems in terms of clusters of cameras and associated images.
Specifically, our goal of camera clustering is to find camera clusters such that all the SfM operations of each cluster can be fitted into a single computer for efficient processing (\textbf{size} constraint) and that all the clusters have sufficient overlapping cameras with adjacent clusters to guarantee a complete reconstruction when their corresponding partial reconstructions are merged together in motion averaging (\textbf{completeness} constraint).

\vspace{-3mm}
\subsubsection{Clustering Formulation}
\vspace{-1mm}
In order to encode the relationships between all the cameras and associated tracks, we introduce a camera graph $\mathcal{G}=\{\mathcal{V},\mathcal{E}\}$,
in which each node $V_i\in\mathcal{V}$ represents a camera $C_i\in\mathcal{C}$,
each edge $e_{ij}\in\mathcal{E}$ with weight $w(e_{ij})$ connects two different cameras $C_i$ and $C_j$.
In the subsequent scalable SfM, both local incremental SfM and bundle adjustment~\cite{eriksson2016} encourage cameras with great numbers of common features to be grouped together for a robust geometry estimation.
We therefore define the edge weight $w(e_{ij})$ as the number of feature correspondences, namely $w(e_{ij}) = |\mathcal{M}_{ij}|$.
Our target is then to partition all the cameras denoted by a graph $\mathcal{G}=\{\mathcal{V},\mathcal{E}\}$ into a set of camera clusters denoted by $\{\mathcal{G}_k|\mathcal{G}_k=\{\mathcal{V}_k,\mathcal{E}_k\}\}$ while satisfying the following \textbf{size} and \textbf{completeness} constraints.

\begin{figure}[t!]
\centering
\includegraphics[width=1.0\linewidth]{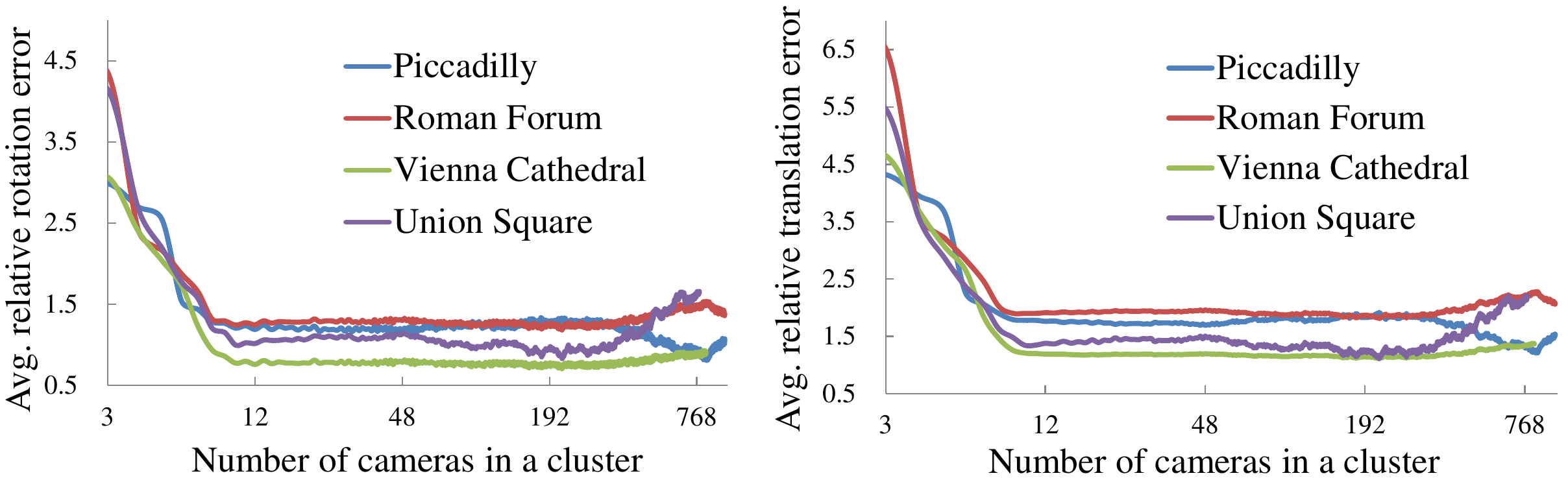}
\caption{
The average relative rotation and translation errors compared with the ground-truth data for different choices of the number of cameras in a cluster for the four Internet data-sets~\cite{wilson2014}.
}
\label{fig:cluster_size_motion_error}
\vspace{-6mm}
\end{figure}

\paragraph{Size constraint}\vspace{-4mm}
We encourage the number of cameras of each camera cluster to be small and of similar size.
First, each camera cluster should be small enough to be fit into a single computer for efficient local SfM operations.
Particularly for local incremental SfM, a comparatively small-scale problem can effectively avoid redundant time-consuming intermediate bundle adjustment~\cite{triggs1999} and possible drifting.
Second, a balanced problem partition stimulates a fully utilization of the distributed computing system.
The size constraint is therefore defined as
\begin{equation}
\begin{aligned}
&\forall{\mathcal{G}_i\in\{\mathcal{G}_k\}},\; |\mathcal{V}_i| \leq \Delta_{\text{up}}\\
&\forall{\mathcal{G}_i,\mathcal{G}_j}\in \{\mathcal{G}_k\},\; |\mathcal{V}_i| \simeq |\mathcal{V}_j|
\end{aligned}
\end{equation}
where $\Delta_{\text{up}}$ is the upper bound of the number of cameras of a cluster.
We can see from Figure~\ref{fig:cluster_size_motion_error} that both the average relative rotation and translation errors computed from local incremental SfM in a cluster first remarkably decrease and then stabilize as the number of cameras in a cluster increases.
The acceptable number of cameras in a cluster is therefore in a large range and we choose $\Delta_{\text{up}}=100$ for the trade-off between accuracy and efficiency.

\begin{figure}[t!]
\centering
\includegraphics[width=1.0\linewidth]{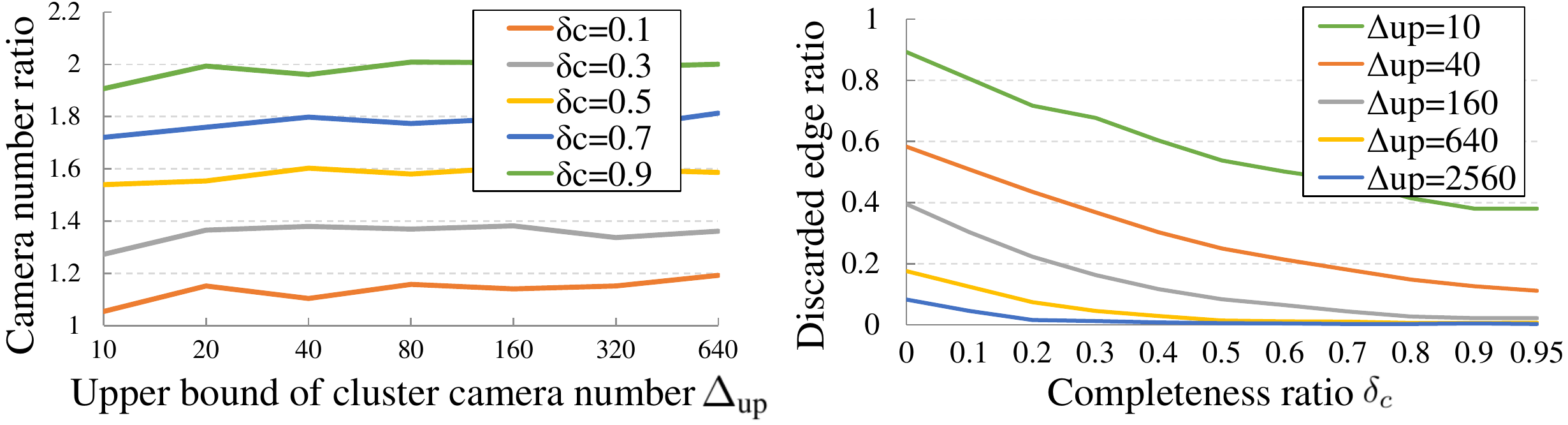}
\caption{Left: the ratio of the total number of cameras of all the clusters to the original number of cameras given different upper bounds of cluster camera numbers $\Delta_\text{up}$ and completeness ratio $\delta_{c}$.
Right: the ratio of discarded edges given different $\Delta_\text{up}$ and $\delta_{c}$.
The plot is based on the statistics of the city-scale data-sets.
}
\label{fig:upper_bound_completeness_ratio}
\vspace{-6mm}
\end{figure}

\paragraph{Completeness constraint}\vspace{-4mm}
The completeness constraint is introduced to preserve camera-to-camera connectivities, which provides relative poses for motion averaging to generate global camera poses.
However, a complete preserving of camera-to-camera connectivities introduces many repeated cameras in different clusters and the size constraint can hardly be satisfied~\cite{bourse2014}.
We therefore define the completeness ratio of a camera cluster $\mathcal{G}_i$ as $\delta(\mathcal{G}_i) = \frac{\sum_{{i \neq j}}{|\mathcal{V}_i \cap \mathcal{V}_j}|}{|\mathcal{V}_i|}$ which quantifies the degree cameras covered in one camera cluster $\mathcal{G}_i$ are also covered by other camera clusters. 
It limits the number of repeated cameras and guarantees that all the clusters have sufficient overlapping cameras with adjacent clusters for a complete reconstruction.
Then, we have 
\begin{equation}
\forall{\mathcal{G}_i} \in \{\mathcal{G}_k\},\;\;\delta(\mathcal{G}_i)\geq\delta_c.
\end{equation}
As shown in Figure~\ref{fig:upper_bound_completeness_ratio}, 
a large completeness ratio $\delta_c$ encourages less loss of camera-to-camera connectivities while results in more duplicated cameras in different clusters.
Balancing the trade-off between accuracy and efficiency,
we choose $\delta_c=0.7$.
Here, less than $5\%$ of camera-to-camera connectivities are discarded and approximately 1.8 times of the original number of cameras are reconstructed in local SfM.
In contrast, exclusive camera clustering ($\delta_c=0$) leads to a loss of $40\%$ of camera-to-camera connectivities.

\vspace{-2mm}
\subsubsection{Clustering Algorithm}
\begin{figure}[t!]
\centering
\includegraphics[width=1.0\linewidth]{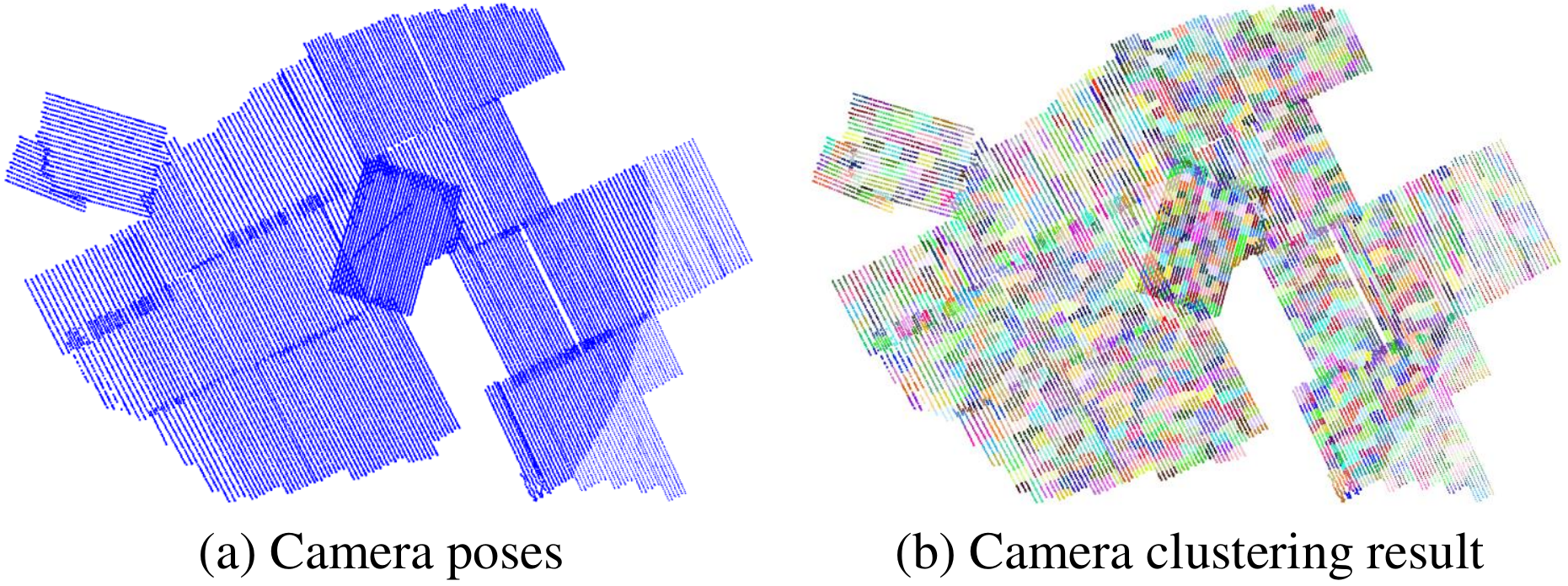}
\caption{
The visual results of our camera clustering algorithm before graph expansion on the City-B data-set.
}
\label{fig:camera_clustering}
\vspace{-6mm}
\end{figure}

\vspace{-2mm}
We propose a two-step algorithm to solve the camera clustering problem.
A sample output of this algorithm is illustrated in Figure~\ref{fig:camera_clustering}.

\paragraph{1. Graph division}\vspace{-4mm}
We guarantee the size constraint by recursively splitting a camera cluster violating the size constraint into smaller components.
Starting with the camera graph $\mathcal{G}$,
we iteratively apply normalized-cut algorithm~\cite{dhillon2007}, which guarantees an unbiased vertex partition, 
to divide any sub-graph $\mathcal{G}_i$ not satisfying the size constraint into two balanced sub-graphs $\mathcal{G}_{i_1}$ and $\mathcal{G}_{i_2}$,
until that no sub-graphs violate the size constraint.
Intuitively, camera pairs with great numbers of common features have high edge weights and are less likely to be cut.

\paragraph{2. Graph expansion}\vspace{-4mm}
We enforce the completeness constraint by introducing sufficient overlapping cameras between adjacent camera clusters.
More specifically, we first sort $\mathcal{E}_\text{dis}$ the edges discarded in graph division by edge weight $w(e_{ij})$ in descending order,
and iteratively add the edge $e_{ij}$ and associate vertices $V_i$ and $V_j$ randomly to one of its connected sub-graphs $\mathcal{G}(V_i)$ and $\mathcal{G}(V_j)$ if the completeness ratio of the subgraph is smaller than $\delta_c$.
Here, $\mathcal{G}(V_i)$ denotes the sub-graph containing vertex $V_i$. 
Such process is iterated until no additional edges can be added to any of the sub-graph.
It is noteworthy that the completeness constraint is not difficult to satisfy after adding a small subset of discarded edges and associated vertices.

The size constraint may be violated after graph expansion, 
and we iterate between graph division and graph expansion until both constraints are satisfied.

\subsection{Camera Cluster Categorization}
The camera clusters from the clustering algorithm are divided into two categories, namely \textbf{independent} and \textbf{interdependent} camera clusters.
We define the final camera clusters from our clustering algorithm as interdependent camera clusters since they share overlapping cameras with adjacent clusters.
Such interdependent clusters are used in subsequent parallel local incremental SfM.
Accordingly, we define all the fully exclusive camera clusters before graph expansion as independent camera clusters which are used in the following parallel 3D point triangulation and parallel bundle adjustment.
We also leverage the independent camera clusters to build a hierarchical camera cluster tree $\mathcal{T}_c$, in which each leaf node corresponds to an independent camera cluster and each non-leaf node is associated with an intermediate camera cluster during the recursive binary graph division.
The hierarchical camera cluster tree is an important structure in the subsequent parallel track generation.
Next, we can base on the camera clusters from our clustering algorithm to implement a scalable SfM pipeline.

\setlength{\textfloatsep}{8pt}
\begin{algorithm}[t!]
\caption{Graph-based camera clustering algorithm} \label{algo:graph_division}
{\scriptsize
\begin{algorithmic}
\State \textbf{Input:} $\mathcal{G}=\{\mathcal{E},\mathcal{V}\}$
\State \textbf{Output:} $\mathbb{G}_\text{out}=\{\mathcal{G}_k|\mathcal{G}_k=\{\mathcal{E}_k,\mathcal{V}_k\}\}$
\State $\mathbb{G}_\text{in} \gets \{\mathcal{G}\}, \mathbb{G}_\text{out} \gets \emptyset$
\While{$\mathbb{G}_\text{in}\neq\emptyset$}
\Comment{\textbf{Iteration between graph division and expansion}}
\State $\mathbb{G}_\text{size} \gets \emptyset$
\Comment{\textbf{Graph division}}
\While{$\mathbb{G}_\text{in}\neq\emptyset$}
\State Choose $\mathcal{G}_i=\{\mathcal{V}_i,\mathcal{E}_i\}$ from $\mathbb{G}_\text{in}$
\State $\mathbb{G}_\text{in}\gets\mathbb{G}_\text{in}-\{\mathcal{G}_i\}$
\If{$|\mathcal{V}_i| \leq \Delta_\text{up}$}
\State $\mathbb{G}_\text{size}\gets\mathbb{G}_\text{size}+\{\mathcal{G}_i\}$
\Else
\State Divide $\mathcal{G}_i$ into $\mathcal{G}_{i_1}$ and $\mathcal{G}_{i_2}$ by normalized-cut~\cite{dhillon2007}
\State $\mathbb{G}_\text{in}\gets\mathbb{G}_\text{in}+\{\mathcal{G}_{i_1}\}+\{\mathcal{G}_{i_2}\}$
\EndIf
\EndWhile

\State$\mathcal{E}_\text{dis}\gets$ edges discarded in graph division 
\Comment{\textbf{Graph expansion}}
\For{each $e_{ij}\in\mathcal{E}_\text{dis}$ sorted by $w(e_{ij})$ in descending order}
\State Select one from $\mathcal{G}(V_i),\;\mathcal{G}(V_j)\in\mathbb{G}_\text{size}$ such that $\delta(\mathcal{G}(V_i))<\delta_c$, $\delta(\mathcal{G}(V_j))\!\!<\!\!\delta_c$ uniformly at random, where $\mathcal{G}(V_i)$ is the sub-graph containing $V_i$ and $\delta(\mathcal{G})$ measures the completeness ratio of $\mathcal{G}$
\If{$\mathcal{G}(V_i)$ is selected}
\State Add $e_{ij}$ and $V_j$ to $\mathcal{G}(V_i)$
\ElsIf{$\mathcal{G}(V_j)$ is selected}
\State Add $e_{ij}$ and $V_i$ to $\mathcal{G}(V_j)$
\EndIf
\EndFor

\For{each $\mathcal{G}_{i}=\{\mathcal{E}_i,\mathcal{V}_i\}\in\mathbb{G}_\text{size}$}
\If{$|\mathcal{V}_i| \leq \Delta_\text{up}$}
\State $\mathbb{G}_\text{out}\gets\mathbb{G}_\text{out}+\{\mathcal{G}_i\}$
\Else
\State $\mathbb{G}_\text{in}\gets\mathbb{G}_\text{in}+\{\mathcal{G}_i\}$
\EndIf
\EndFor
\EndWhile
\end{algorithmic}
}
\end{algorithm}

\section{Scalable Implementation}
\subsection{Track Generation}\label{sec:track_generation}
The first step of scalable SfM is to use the pair-wise feature correspondences to generate globally consistent tracks across all the images, 
and the problem is solved by a standard Union-Find~\cite{moulon2012} algorithm.
However, as the size of the input images scales up, it gradually becomes impossible to concurrently load all the feature and associate match files into the memory of a single computer for track generation.
We therefore base on the hierarchical camera cluster tree $\mathcal{T}_c$ to perform track generation and avoid caching all the features and correspondences into memory at once.
In detail, we define $N_i^k$ as the node in the $k$th level of $\mathcal{T}_c$,
and $N_{i_1}^{k+1}$ and $N_{i_2}^{k+1}$ are respectively the left and right child of $N_i^k$.
For the track generation sub-problem associated with sibling leaf nodes ${N}^{k+1}_{i_1}$ and ${N}^{k+1}_{i_2}$, we load all their features and correspondences into memory, generate the tracks corresponding to ${N}_i^k$, reallocate the memory of features and correspondences, and save the tracks associated with ${N}^{k}_{i}$ into storage.
As for the two sibling non-leaf nodes $N^{l+1}_{j_1}$ and $N^{l+1}_{j_2}$, we only load the correspondences and tracks associated with both nodes, merge them, and save the tracks corresponding to $N^{l}_{j}$ into storage.
Such processes are iteratively performed from the bottom up until the globally consistent tracks with respect to the root node of $\mathcal{T}_c$ are obtained.
All the track generation processes associated with each level of $\mathcal{T}_c$ are handled in parallel under a standard framework of MapRedeuce~\cite{dean2008}.

\begin{figure}[t!]
\centering
\includegraphics[width=1.0\linewidth]{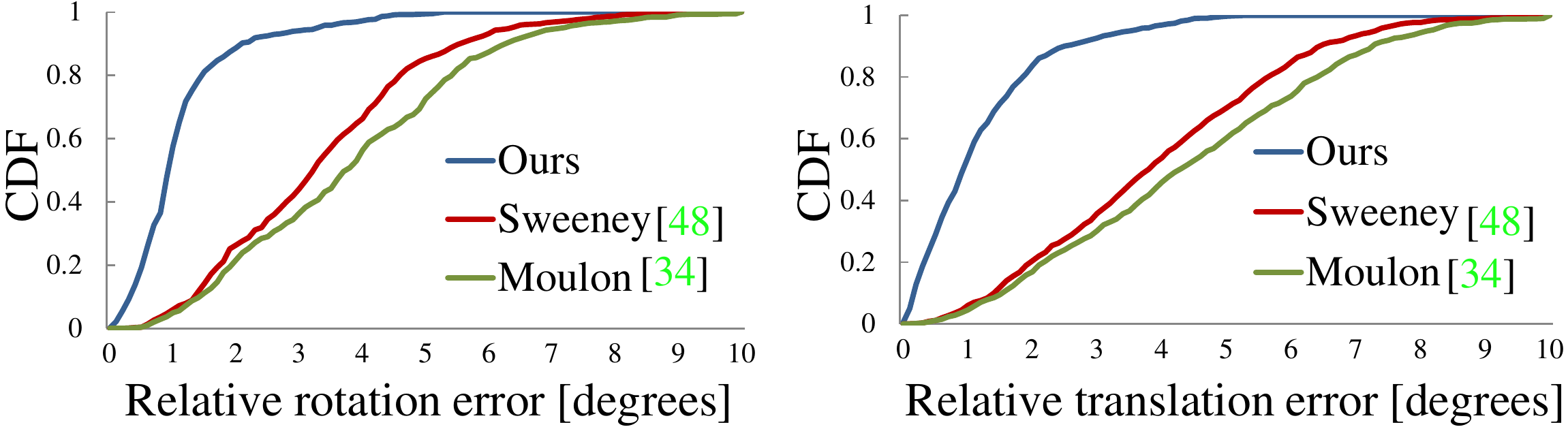}
\caption{
The comparison of the accuracy of the generated relative rotations and relative translations between the motion averaging methods~\cite{moulon2013,theia-manual} and our approach.
The cumulative distribution functions (CDF) are based on the Internet data-sets~\cite{wilson2014}. 
}
\label{fig:relative_error}
\vspace{-1mm}
\end{figure}

\subsection{Local Incremental SfM}\label{sec:local_sfm}
For the cameras and corresponding tracks of every interdependent camera cluster $\mathcal{C}^k$ denoted by the sub-graph $\mathcal{G}_k\!=\!\{\mathcal{E}_k,\mathcal{V}_k\}$, we perform local incremental SfM in parallel.
Local incremental SfM is vital to the subsequent motion averaging in two aspects.
First, RANSAC~\cite{fischler1981} based filters and repeated partial bundle adjustment~\cite{triggs1999} can remove erroneous epipolar geometry and feature correspondences.
Second, incremental SfM considers robust $N$-view ($\geq3$) pose estimation~\cite{kneip2011,nister2004efficient} and produces superior accurate and robust relative rotations and translations than the generally adopted essential matrix based~\cite{arie-nachimson2012,brand2004,govindu2001,ozyesil2015} and trifocal tensor based methods~\cite{jiang2013,moulon2013} even for the camera pairs with weak association, large angle of views, and great scale variation.
Figure~\ref{fig:relative_error} and the statistics of the benchmark data-sets~\cite{strecha2008benchmarking} ($\delta \bar{R}$ and $\delta \bar{t}$) in Table~\ref{tab:benchmark} confirm the statement above.

\subsection{Motion Averaging}\label{sec:motion_averaging}
Now, all the relative motions of camera pairs with feature correspondences from local incremental SfM are used to compute the global camera poses.
The work in~\cite{chatterjee2013} is first adopted for efficient and robust global rotation averaging.

\vspace{-3mm}
\subsubsection{Translation Averaging}\label{sec:translation_averaging}
\vspace{-1mm}
Translation averaging is challenging for two reasons.
First, it is difficult to discard erroneous epipolar geometry resulted from noisy feature correspondences.
Second, an essential matrix can only encode the direction of a relative translation~\cite{ozyesil2015}.
Thanks to local incremental SfM, 
the majority of erroneous epipolar geometry is filtered,
and the only problem remained is to solve the scale ambiguity.

The work in~\cite{cui2015} first globally averages the scales of all the relative translations and perform a convex $\ell1$ optimization to solve scale-aware translation averaging.
{\"{O}}zyesil~\etal~\cite{ozyesil2015} obtain the convex ``least unsquared deviations" formulation by introducing a complicated quadratic constraint.
Given that all the relative translations $\{\mathbf{t}_{ij}^k\}$ from one camera cluster $\mathcal{C}_k$ are up to the same scale factor $\alpha_k$,
we instead formulate our translation averaging as a convex $\ell1$ problem by solving the camera positions and cluster scales simultaneously.
Obviously, the scale factors computed in terms of clusters are more robust than the pair-wise scales~\cite{cui2015,ozyesil2015} in terms of relative poses,
especially for the camera pairs with weak association.

\begin{figure}[t!]
\centering
\includegraphics[width=1.0\linewidth]{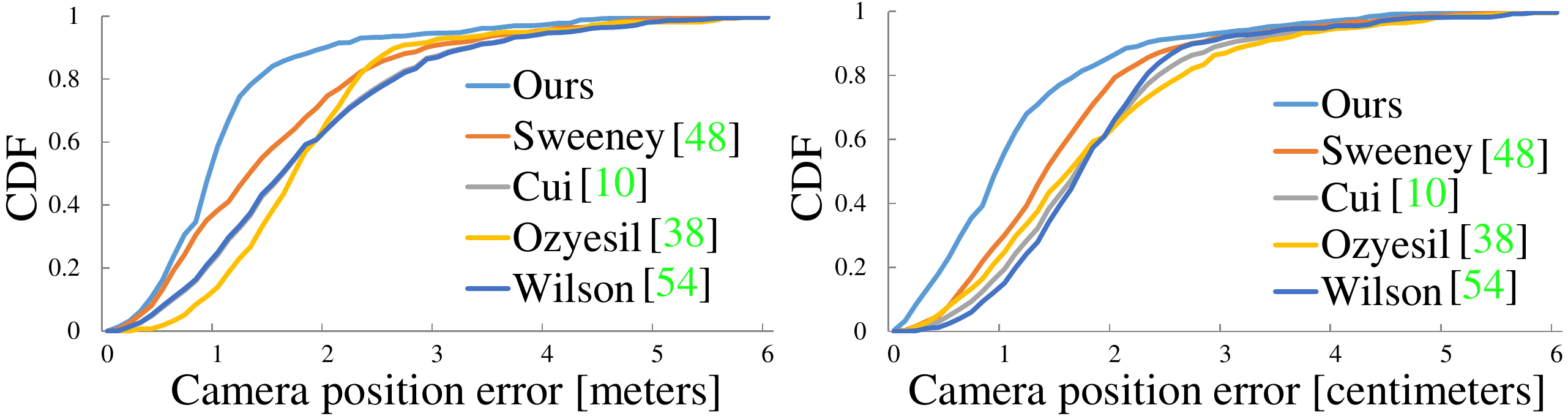}
\caption{The comparison of the camera position errors with the state-of-the-art translation averaging methods~\cite{cui2015,ozyesil2015,theia-manual,wilson2014} on both benchmark~\cite{strecha2008benchmarking} (left) and Internet~\cite{wilson2014} (right) data-sets given the same input global rotations from~\cite{chatterjee2013} and relative translations from local incremental SfM.}
\label{fig:position_error}
\vspace{-1mm}
\end{figure}


With the global rotations $\{\mathbf{R}_i\}$ computed from~\cite{chatterjee2013} fixed,
a linear equation of camera positions can be obtained as:
\vspace{-1mm}
\begin{equation}\label{eq:translation_averaging}
\alpha_k\mathbf{t}_{ij}^k = \mathbf{R}_j(\mathbf{c}_i-\mathbf{c}_j),
\vspace{-1mm}
\end{equation}
where $\mathbf{t}^k_{ij}$ is a relative translation between two cameras $C_i$ and $C_j$ estimated in the $k$th cluster associated with a scale $\alpha_k$.
Equation~\ref{eq:translation_averaging} can be rewritten as:
$\alpha_k\mathbf{R}_j^T\mathbf{t}_{ij}^k = \mathbf{c}_i-\mathbf{c}_j$.
Then we form the representations of all the cluster scales and camera positions as $\mathbf{x}_{s}=[\alpha_1,\cdot\cdot\cdot,\alpha_M]^T$ and $\mathbf{y}_{c} = [\mathbf{c}_1,\cdot\cdot\cdot,\mathbf{c}_N]^T$ respectively,
and we have:
\vspace{-1mm}
\begin{equation}
\underbrace{[\;\cdot\cdot\cdot \mathbf{p} \cdot\cdot\;\cdot\;]}_{\mathbf{A}^{k}_{ij}}\mathbf{x}_{s} = \underbrace{[\;\cdot\cdot\cdot \mathbf{I} \cdot\cdot\cdot -\mathbf{I} \cdot\cdot\;\cdot\;]}_{\mathbf{B}_{ij}}\mathbf{y}_{c}.
\vspace{-1mm}
\end{equation}
Here, $\mathbf{A}^{k}_{ij}$ is a $3 \times M$ matrix with an appropriate location of $k$ replaced by $\mathbf{p}=\mathbf{R}_j^T\mathbf{t}_{ij}^k$, and $\mathbf{0}_{3 \times 1}$ otherwise.
$\mathbf{B}_{ij}$ is a $3\times 3N$ matrix with appropriate locations of $i$ and $j$ replaced by $\mathbf{I}_{3\times3}$ and $-\mathbf{I}_{3\times3}$ respectively,
and $\mathbf{0}_{3\times3}$ otherwise.
Then, we can collect all such linear equations from the available camera-to-camera connectivities into the following single linear equation system:
\vspace{-1mm}
\begin{equation}\label{eq:translation_linear_equation}
\mathbf{A}\mathbf{x}_s=\mathbf{B}\mathbf{y}_c,
\vspace{-1mm}
\end{equation}
where $\mathbf{A}$ and $\mathbf{B}$ are sparse matrices made by stacking all the associate matrices $\mathbf{A}^{k}_{ij}$ and $\mathbf{B}_{ij}$ respectively. 

\begin{figure}[t!]
\begin{subfigure}{0.33\linewidth}
  \includegraphics[width=1.0\linewidth]{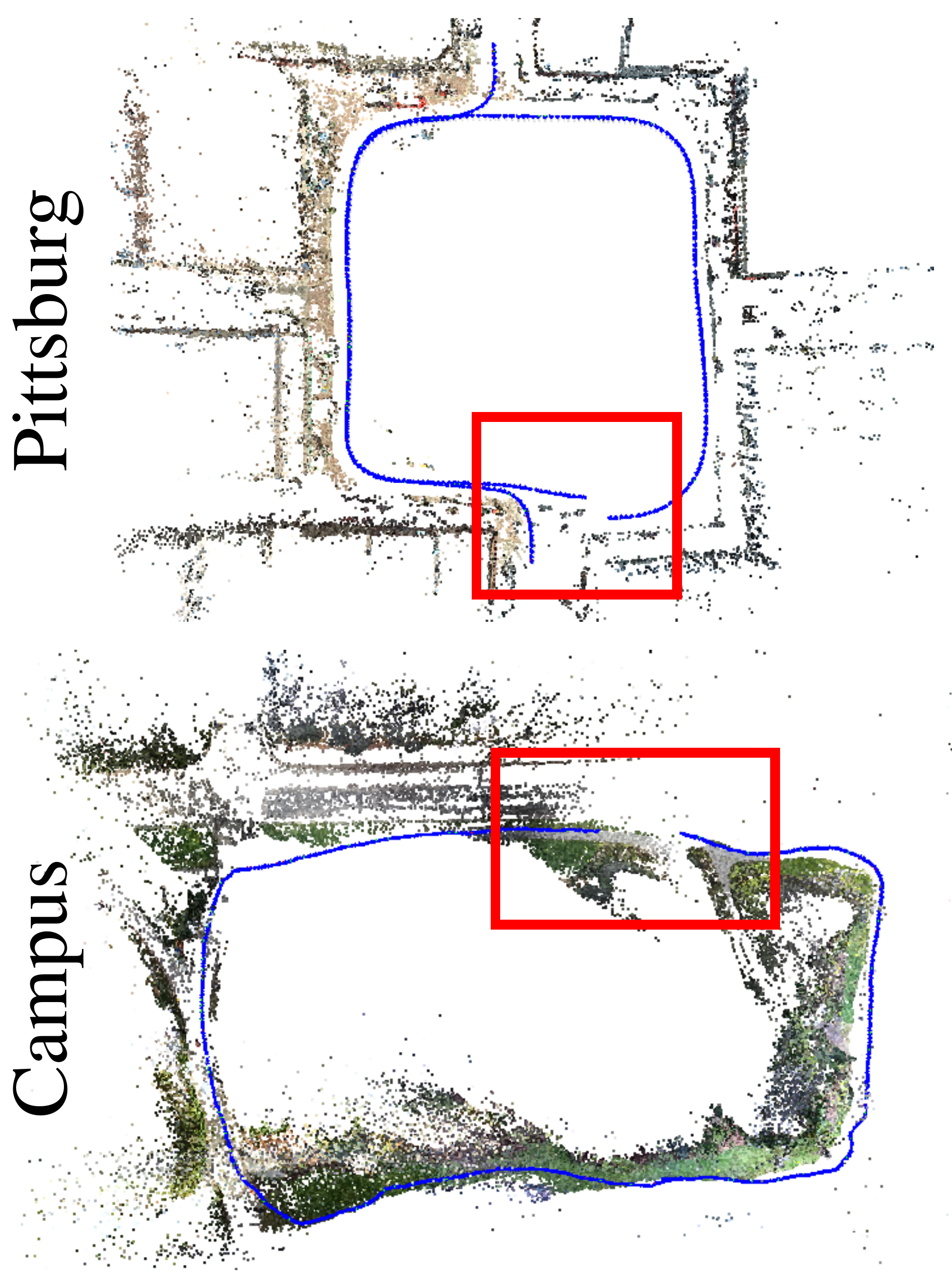}
  \caption{Bhowmick~\cite{bhowmick2014}}
\end{subfigure}
\begin{subfigure}{0.31\linewidth}
  \includegraphics[width=1.0\linewidth]{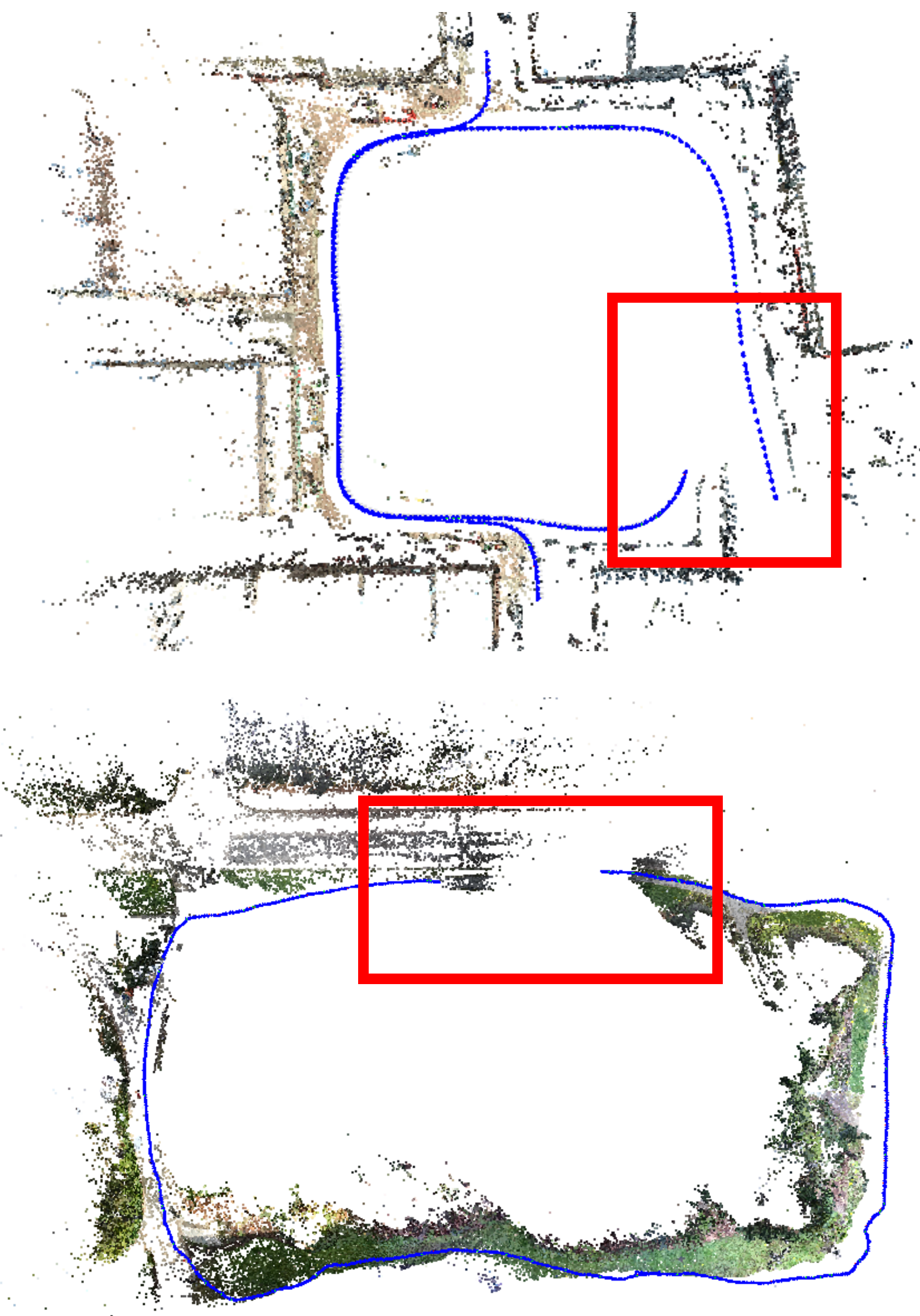}
  \caption{Sweeney~\cite{sweeney2017}}
\end{subfigure}
\begin{subfigure}{0.30\linewidth}
  \includegraphics[width=1.0\linewidth]{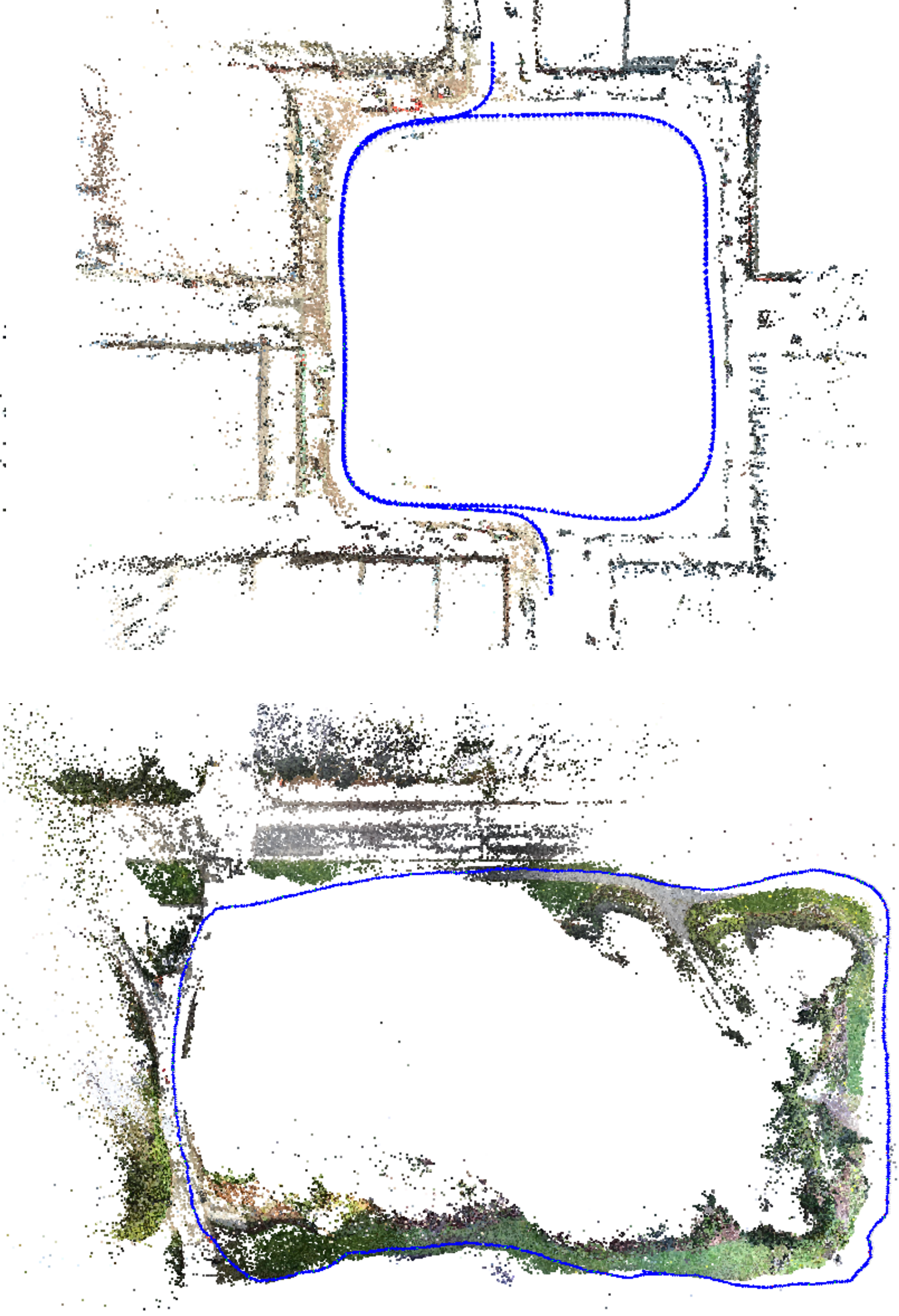}
  \caption{Ours}
\end{subfigure}
~
\caption{The visual comparison between the approaches of exclusive camera clusters~\cite{bhowmick2014,sweeney2017} and our approach using interdependent camera clusters on the sequential data-sets with close-loop~\cite{cui2015}.
}
\label{fig:closeloop}
\vspace{-2mm}
\end{figure}

\begin{table*}[t!]
\begin{center}
\resizebox{\linewidth}{!}
{
\begin{tabular}{c|c|c|c|c|c|c|>{\centering}p{1.1cm}|c|c|c|c|c|>{\centering}p{1.1cm}|c|c|c|c|c|>{\centering}p{1.1cm}}
\hline 
\multirow{3}{*}{Data-set} & \multirow{3}{*}{\# images} & \multicolumn{6}{c|}{Average epipolar error {[}pixels{]}} & \multicolumn{6}{c|}{Number of connected camera pairs} & \multicolumn{6}{c}{Number of 3D points}\tabularnewline
\cline{3-20} 
 &  & \multirow{2}{*}{\cite{bhowmick2014}} & \multirow{2}{*}{\cite{sweeney2017}} & \multicolumn{4}{c|}{Ours} & \multirow{2}{*}{\cite{sweeney2017}} & \multirow{2}{*}{\cite{bhowmick2014}} & \multicolumn{4}{c|}{Ours} & \multirow{2}{*}{\cite{bhowmick2014}} & \multirow{2}{*}{\cite{sweeney2017}} & \multicolumn{4}{c}{Ours}\tabularnewline
 &  &  &  & $\delta_{c}\!=\!0.1$ & $\delta_{c}\!=\!0.3$  & $\delta_{c}\!=\!0.5$  & $\delta_{c}\!=\!0.7$  &  &  & $\delta_{c}\!=\!0.1$ & $\delta_{c}\!=\!0.3$  & $\delta_{c}\!=\!0.5$  & $\delta_{c}\!=\!0.7$  &  &  & $\delta_{c}\!=\!0.1$ & $\delta_{c}\!=\!0.3$  & $\delta_{c}\!=\!0.5$  & $\delta_{c}\!=\!0.7$ \tabularnewline
\hline
Pittsburg & 388 & 6.74 & 5.48 & 2.88 & 0.90 & 0.81 & 0.78 & 7.4K & 6.7K & 8.2K & 9.4K & 10.1K & 10.3K & 57K & 64K & 74K & 77K & 80K & 81K\tabularnewline
\hline 
Campus & 1550 & 3.42 & 3.93 & 2.74 & 1.22 & 0.72 & 
0.66
 & 43.2K & 34.7K & 61.7K & 69.4K & 72.2K & 76.0K & 156K & 173K & 248K & 252K & 276K & 294K\tabularnewline
\hline 
\end{tabular}
}
\end{center}
\vspace{-3mm}
\caption
{
The statistical comparison between the methods of exclusive camera clusters~\cite{bhowmick2014,sweeney2017} and our work using interdependent camera clusters on the sequential data-sets~\cite{cui2015}.
The statistics of our method with different completeness ratios $\delta_c$ are also provided to verify the effectiveness of the completeness constraint.
}
\label{tab:close_loop}
\vspace{-6mm}
\end{table*}

After removing the gauge freedom by setting $\mathbf{c}_1=\mathbf{0}_{3\times 1}$ and $\alpha_1=1$,
we can obtain the positions of all the cameras by solving the following robust convex $\ell1$ optimization problem that is more robust to outliers than $\ell2$ methods and converges rapidly to a global optimum,
\vspace{-1mm}
\begin{equation}\label{eq:translation_averaging_sum}
\arg\min_{\mathbf{x}_s,\mathbf{y}_c}{||\mathbf{A}\mathbf{x}_s-\mathbf{B}\mathbf{y}_c||_{1}}.
\vspace{-1mm}
\end{equation}
Since the baseline length is encoded by the changes of cluster scales,
our translation averaging algorithm can effectively handle the scale ambiguity,
especially for collinear camera motion,
and is much well-posed than the essential matrix based approaches~\cite{brand2004,govindu2001,ozyesil2015,wilson2014},
which only consider the directions of relative translations and are limited to the parallel rigid graph~\cite{ozyesil2015}.

\subsection{Bundle Adjustment}\label{sec:bundle_adjustment}

For each independent camera cluster, we triangulate~\cite{hartley2003multiple} their corresponding 3D points with sufficient visible cameras ($\geq 3$) from their feature correspondences validated by local incremental SfM based on the averaged global camera geometry.
Then, we follow the state-of-the-art algorithm proposed by Eriksson~\etal~\cite{eriksson2016} for distributed bundle adjustment.
Since this work~\cite{eriksson2016} declares to have no restriction on the partitions of cameras,
we refer to the independent camera clusters with their associate cameras, tracks and projections as the sub-problems of the objective function of bundle adjustment.

\vspace{-1mm}
\subsection{Discussion}
\label{sec:motion_averaging_discussion}
\vspace{-1mm}
Given the same global camera rotations from~\cite{chatterjee2013} and relative translations from local SfM,
Figure~\ref{fig:position_error} verifies that our translation averaging algorithm recovers more accurate camera positions than the state-of-the-art translation averaging methods~\cite{cui2015,moulon2013,ozyesil2015,theia-manual,wilson2014}.
Although the optimal solution to no loss of relative motions compared with the original camera graph can hardly be obtained in our clustering algorithm,
the statistical comparison shown in Table~\ref{tab:benchmark} still demonstrates the superior accuracy of camera poses from our pipeline
over the state-of-the-art SfM approaches~\cite{cui2015,moulon2013,theia-manual,wu2013} on the benchmark data-set~\cite{strecha2008benchmarking}.

Figure~\ref{fig:closeloop} shows the comparison with the hybrid SfM methods~\cite{bhowmick2014,sweeney2017} using exclusive camera clusters on the data-sets~\cite{cui2015} consisted of sequential images with close-loop.
We regard our independent camera clusters as the clusters adopted in~\cite{bhowmick2014,sweeney2017}.
We can see that our global method with interdependent camera clusters successfully guarantees close-loop while those~\cite{bhowmick2014,sweeney2017} with exclusive camera clusters fail.

The statistical comparison with the hybrid SfM methods~\cite{bhowmick2014,sweeney2017} are shown in Table~\ref{tab:close_loop}.
To measure the consistency of camera poses,
we use the epipolar error that is the median distance between the features and corresponding epipolar lines computed from the feature correspondences of all the camera pairs, 
the number of camera pairs connected by 3D points,
and the number of final 3D points.
Since our clustering algorithm introduces sufficient camera connectivities for a fully constrained global motion averaging rather than directly merging exclusive camera clusters~\cite{bhowmick2014,sweeney2017},
the epipolar error of our approach is only $10\%\!-\!20\%$ of that of the work~\cite{bhowmick2014,sweeney2017},
the number of connected camera pairs is $1.8\!-\!4.5$ times of that of the work \cite{bhowmick2014,sweeney2017},
and we generate $1.3\!-\!3.0$ times more 3D points than the work in~\cite{bhowmick2014,sweeney2017}.
Table~\ref{tab:close_loop} also provides the results of our approach with different complements ratio. We can see that a larger completeness ratio, namely more camera-to-camera connectivities, guarantees a more accurate and complete sparse reconstruction.

\begin{table}[t!]
\begin{center}
\resizebox{\linewidth}{!}
{
\begin{tabular}{c||c||c||c|c|c||c|c|c||c|c|c|c}
\multirow{2}{*}{Data-set} & \multicolumn{1}{c||}{Wu~\cite{wu2013}} & \multicolumn{1}{c||}{Cui~\cite{cui2015}} & \multicolumn{3}{c||}{Moulon~\cite{moulon2013}} & \multicolumn{3}{c||}{Sweeney~\cite{theia-manual}} & \multicolumn{4}{c}{Ours}\tabularnewline
 & $\bar{x}_{BA}$  & $\bar{x}$  & $\delta\bar{R}$  & $\delta\bar{t}$  & $\bar{x}$  & $\delta\bar{R}$  & $\delta\bar{t}$  & $\bar{x}$  & $N_{s}$  & $\delta\bar{R}$  & $\delta\bar{t}$  & $\bar{x}$ \tabularnewline
\hline 
FountainP11 & 7.7 & 16.2 & 0.08 & 0.07 & 69.7 & 0.11 & 0.13 & 11.3 & 2 & \textbf{0.04} & \textbf{0.05} & \textbf{2.9}\tabularnewline
\hline 
EntryP10 & 8.5 & 59.4 & 0.13 & 0.16 & 71.5 & 0.11 & 0.19 & 67.1 & 2 & \textbf{0.04} & \textbf{0.08} & \textbf{6.0}\tabularnewline
\hline 
HerzJesuP8 & 10.7 & 21.7 & 0.69 & 4.26 & 69.7 & \textbf{0.68} & \textbf{4.23} & 5.7 & 2 & 0.71 & 4.58 & \textbf{3.7}\tabularnewline
\hline 
HerzJesuP25 & 21.3 & 73.1 & 0.11 & 0.19 & 293.0 & 0.13 & 0.20 & \textbf{8.6} & 4 & \textbf{0.03} & \textbf{0.07} & 17.3\tabularnewline
\hline 
CastleP19 & 320.1 & 573.9 & 0.34 & 0.64 & 544.6 & 0.43 & 0.76 & 619.1 & 3 & \textbf{0.07} & \textbf{0.08} & \textbf{23.3}\tabularnewline
\hline 
CastleP30 & 204.1 & 671.2 & 0.34 & 0.64 & 739.6 & 0.32 & 0.59 & 566.8 & 5 & \textbf{0.06} & \textbf{0.09} & \textbf{35.8}\tabularnewline
\hline 
\end{tabular}
}
\end{center}
\vspace{-2mm}
\caption
{
The comparison with the global SfM methods~\cite{cui2015,moulon2013,theia-manual} on the benchmark data-sets~\cite{strecha2008benchmarking}. 
Specifically, $\bar{x}$ is the average position error of camera optical centers (measured in millimeters) after motion averaging and before bundle adjustment.
Since repeated intermediate bundle adjustment is performed in incremental SfM~\cite{wu2013},
the average position error after bundle adjustment $\bar{x}_{BA}$ is shown for~\cite{wu2013}.
$N_s$ is the number of camera clusters,
$\delta \bar{R}$ is the average relative rotation error measured in degrees and $\delta \bar{t}$ denotes the average relative translation angular error in degrees.
}
\label{tab:benchmark}
\vspace{-1mm}
\end{table}

\begin{table*}[t!]
\begin{center}
\resizebox{\linewidth}{!}
{
\begin{tabular}{c|c||c|c|c|c|c|c|c|c|c|c|c|c||c|c|c|c|c|c|c|c|c|c|c|c|c}
\hline 
\multirow{3}{*}{Datasets} &  & \multicolumn{12}{c||}{Accuracy {[}meters{]}} & \multicolumn{13}{c}{Time {[}seconds{]}}\tabularnewline
\cline{3-27} 
 & \# images & \multicolumn{2}{c|}{1DSfM~\cite{wilson2014}} & \multicolumn{2}{c|}{Colmap~\cite{schoenberger2016sfm}} & \multicolumn{2}{c|}{Cui~\cite{cui2015}} & \multicolumn{2}{c|}{Sweeney~\cite{sweeney2016}} & \multicolumn{2}{c|}{Theia~\cite{theia-manual}} & \multicolumn{2}{c||}{Ours} & \multicolumn{2}{c|}{1DSfM~\cite{wilson2014}} & \multicolumn{2}{c|}{Colmap~\cite{schoenberger2016sfm}} & \multicolumn{2}{c|}{Cui~\cite{cui2015}} & \multicolumn{2}{c|}{Sweeney~\cite{sweeney2016}} & \multicolumn{2}{c|}{Theia~\cite{theia-manual}} & \multicolumn{3}{c}{Ours}\tabularnewline
 &  & $N_{c}$  & $\tilde{x}_{BA}$  & $N_{c}$  & $\tilde{x}_{BA}$  & $N_{c}$  & $\tilde{x}_{BA}$  & $N_{c}$  & $\tilde{x}_{BA}$  & $N_{c}$  & $\tilde{x}_{BA}$  & $N_{c}$  & $\tilde{x}_{BA}$  & $T_{BA}$  & $T_{\text{\ensuremath{\sum}}}$  & $T_{BA}$  & $T_{\text{\ensuremath{\sum}}}$  & $T_{BA}$  & $T_{\text{\ensuremath{\sum}}}$  & $T_{BA}$  & $T_{\text{\ensuremath{\sum}}}$  & $T_{BA}$  & $T_{\text{\ensuremath{\sum}}}$  & $T_{LS}$  & $T_{BA}$  & $T_{\text{\ensuremath{\sum}}}$ \tabularnewline
\hline 
Alamo  & 577 & 529 & 0.3 & 552 & 0.3 & 540 & 0.5 & 533 & 0.4 & \textbf{558} & 0.4 & 549 & \textbf{0.2} & 752  & 910  & 499 & 840 & 476 & 568 & 129 & 198 & 413 & 497 & 173 & 63 & 264\tabularnewline
\hline 
Ellis Island & 227 & 214 & \textbf{0.3} & 209 & 0.6 & 206 & 0.7 & 203 & 0.5 & 220 & 4.7 & \textbf{221} & 0.5 & 139 &  171 & 137 & 301 & 158 & 209 & 14 & 33 & 14 & 28 & 26 & 14 & 45\tabularnewline
\hline 
Metropolis & 341 & 291 & 0.5 & \textbf{324} & 1.4 & 281 & 3.0 & 272 & 0.4 & 321 & 1.0 & 298 & \textbf{0.2} & 201  & 244 & 302 & 532 & 27 & 64 & 94 & 161 & 34 & 47 & 88 & 26 & 125\tabularnewline
\hline 
Montreal N.D. & 450 & 427 & 0.4 & 437 & \textbf{0.3} & 433 & \textbf{0.3} & 416 & \textbf{0.3} & \textbf{448} & 0.4 & 445 & \textbf{0.3} & 1135  & 1249 & 352 & 688 & 632 & 678 & 133 & 266 & 107 & 164 & 167 & 72 & 261\tabularnewline
\hline 
Notre Dame  & 553 & 507 & 1.9 & 543 & 0.5 & \textbf{547} & \textbf{0.2} & 501 & 1.2 & 540 & \textbf{0.2} & 514 & \textbf{0.2} & 1445  & 1599 & 432 & 708 & 458 & 549 & 161 & 247 & 196 & 331 & 246 & 64 & 338\tabularnewline
\hline 
NYC Library & 332 & 295 & 0.4 & 304 & 0.6 & 303 & \textbf{0.3} & 294 & 0.4 & \textbf{321} & 0.9 & 290 & \textbf{0.3} & 392  & 468 & 311 & 412 & 169 & 210 & 83 & 154 & 47 & 62 & 79 & 52 & 144\tabularnewline
\hline 
Piazza del Popolo & 350 & 308 & 2.2 & 332 & 1.2 & \textbf{336} & 1.4 & 302 & 1.8 & 326 & 1.0 & 334 & \textbf{0.5} & 191  & 249 & 246 & 336 & 126 & 191 & 72 & 101 & 46 & 61 & 72 & 16 & 93\tabularnewline
\hline 
Piccadilly & 2152 & 1956 & 0.7 & 2062 & 0.6 & 1980 & \textbf{0.4} & 1928 & 1.0 & 2055 & 0.7 & \textbf{2114} & \textbf{0.4} & 2425  & 3483 & 623 & 1814 & 984 & 1553 & 702 & 1246 & 72 & 330 & 932 & 542 & 1614\tabularnewline
\hline 
Roman Forum  & 1084 & 989 & \textbf{0.2} & 1062 & 1.7 & 1033 & 2.8 & 966 & 0.7 & 1045 & 2.2 & \textbf{1079} & 0.4 & 1245  & 1457 & 823 & 1122 & 310 & 482 & 847 & 1232 & 183 & 244 & 604 & 201 & 902\tabularnewline
\hline 
Tower of London & 572 & 414 & 1.0 & 450 & \textbf{0.7} & \textbf{458} & 1.2 & 409 & \textbf{0.9} & 456 & 1.4 & \textbf{458} & 1.0 & 606  & 648 & 542 & 665 & 488 & 558 & 92 & 246 & 130 & 154 & 320 & 75 & 410\tabularnewline
\hline 
Union Square & 789 & 710 & 3.4 & \textbf{726} & 2.8 & 570 & 4.2 & 701 & 2.1 & 720 & 5.0 & 720 & \textbf{1.5} & 340  & 452 & 430 & 532 & 45 & 99 & 102 & 243 & 27 & 48 & 145 & 50 & 207\tabularnewline
\hline 
Vienna Cathedral & 836 & 770 & \textbf{0.4} & \textbf{799} & 1.2 & 774 & 1.6 & 771 & 0.6 & 797 & 2.6 & 793 & 0.5 & 2837  & 3139 & 930 & 1254 & 438 & 580 & 422 & 607 & 111 & 244 & 712 & 167 & 905\tabularnewline
\hline 
Yorkminster & 437 & 401 & \textbf{0.1} & \textbf{416} & 0.8 & 407 & 0.6 & 409 & 0.3 & 414 & 1.4 & 407 & 0.3 & 777  & 899 & 724 & 924 & 602 & 662 & 71 & 102 & 59 & 92 & 199 & 67 & 281\tabularnewline
\hline 
\end{tabular}
}
\end{center}
\vspace{-3mm}
\caption
{
The comparison of the Internet data-sets~\cite{wilson2014}.
We regard the SfM results of~\cite{snavely2006} as the reference model.
$N_c$ denotes the number of registered cameras and $\tilde{x}_{BA}$ is the median camera position error in meters after bundle adjustment. 
We also compare the time overhead, and introduce $T_{LS}$, $T_{BA}$ and $T_{\Sigma}$ to denote the local incremental SfM time, full bundle adjustment time, and total running time respectively.
}
\label{tab:internet}
\vspace{-4mm}
\end{table*}

\section{Experiments}
\paragraph{Implementation}
We implement our approach in C++ and perform all the experiments on a distributed computing system consisted of 10 computers each of which has 6-Core (12 threads) Intel 3.40 GHz processors and 128 GB memory.
All the computers are deployed on a scalable network file system similar to Hadoop File System.
We implement a multicore bundle adjustment solver similar to PBA~\cite{wu2011} to solve all the non-linear optimization problems,
and a $\ell1$ solver like~\cite{emmanuel2005} to solve Equation~\ref{eq:translation_averaging_sum}.
We also utilize Graclus~\cite{dhillon2007} to handle the normalized-cut problem.

\paragraph{Benchmark data-sets}\vspace{-5mm}
The statistics of the comparisons of the benchmark data-sets~\cite{strecha2008benchmarking} with absolute measurements of camera poses between the state-of-the-art methods~\cite{cui2015,moulon2013,theia-manual,wu2013} and our proposed method are shown in Table~\ref{tab:benchmark}.
Since the number of cameras of the largest benchmark data-set CastleP30 is only 30,
we set $\Delta_\text{up}=7$ rather than $\Delta_\text{up}=100$ adopted by our pipeline to force that valid camera clusters can be generated.
Specifically, we can see that the average errors of relative rotations ($\delta\bar{R}$), relative translations ($\delta\bar{t}$), and corresponding camera positions ($\bar{x}$) from our algorithm are all obviously smaller than the work in~\cite{cui2015,moulon2013,theia-manual,wu2013}.

\begin{figure}[t!]
\centering
\includegraphics[width=1.0\linewidth]{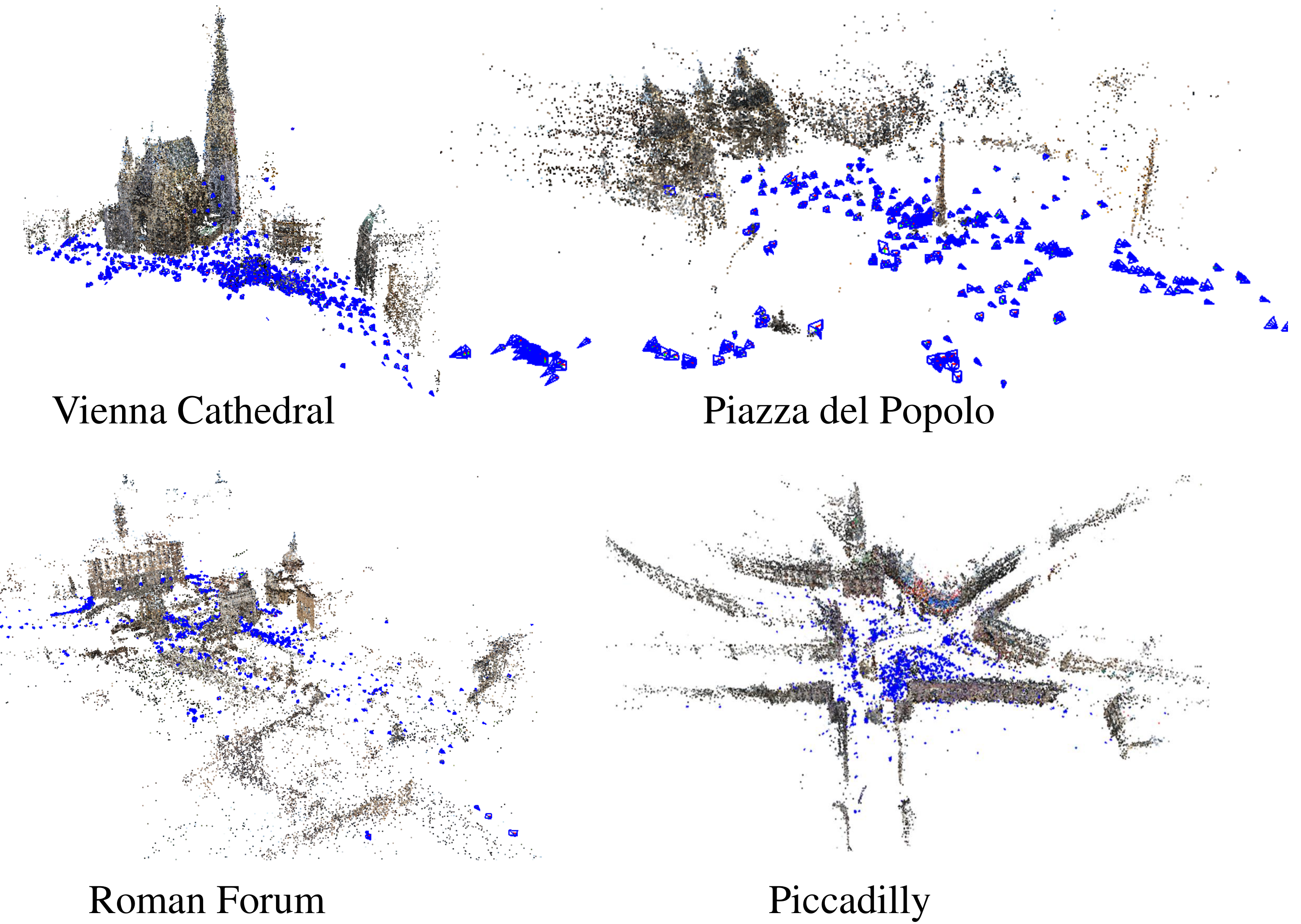}
\caption{
The SfM visual results of the Internet data-sets~\cite{wilson2014}.
}
\label{fig:internet}
\vspace{-1mm}
\end{figure}

\paragraph{Internet data-sets}\vspace{-5mm}
Table~\ref{tab:internet} shows the statistical comparisons with the state-of-the-art SfM pipelines~\cite{cui2015,schoenberger2016sfm,sweeney2016,theia-manual,wilson2014} on the Internet data-set.
We can see that our approach achieves the best accuracy measured by the median camera position errors (in meters) after bundle adjustment in 8 out of 13 data-sets.
Moreover, we register the most cameras in 4 out of 13 data-sets.
Among these methods~\cite{cui2015,schoenberger2016sfm,sweeney2016,theia-manual,wilson2014}, Theia SfM~\cite{theia-manual} is the most efficient.
We can therefore conclude that our SfM pipeline achieve slightly better accuracy and its efficiency is comparable to the state-of-the-art methods~\cite{cui2015,schoenberger2016sfm,sweeney2016,theia-manual,wilson2014} on the data-sets captured in the wild.

\begin{figure*}[t!]
\centering
\includegraphics[width=1.0\linewidth]{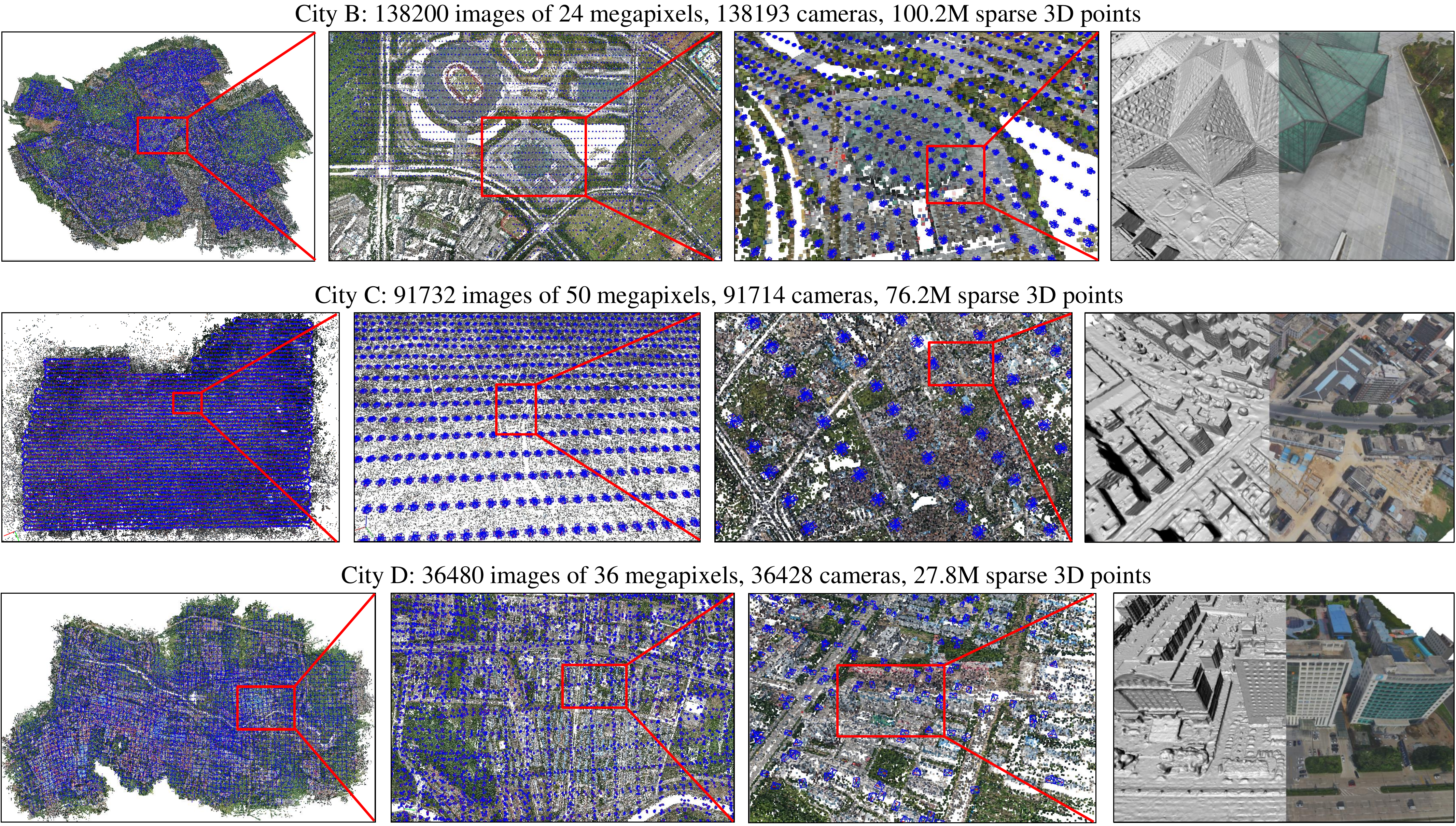}
\caption{
The SfM visual results of the city-scale data-sets.
In each row, the figures from left to right zoom successively closer to the representative buildings.
The delicate 3D model is shown in the last figure to qualitatively demonstrate the accuracy of camera poses.
}
\label{fig:cityscale}\vspace{-3mm}
\end{figure*}

\begin{table*}[t!]
\begin{center}
\resizebox{1.0\linewidth}{!}
{
\begin{tabular}{c|c|c|c|c|c|c|c|c|c|c|c|c|c|c|c|c|c|c}
\hline 
\multirow{3}{*}{Data-set} & \multirow{3}{*}{\# images} & \multirow{3}{*}{Resolution} & \multirow{3}{*}{$\tilde{N}_{f}$} & \multirow{3}{*}{$N_{k}$ } & \multicolumn{3}{c|}{Clustering time {[}minutes{]}} & \multicolumn{5}{c|}{Pipeline time {[}hours{]}} & \multicolumn{6}{c}{Peak memory {[}GB{]}}\tabularnewline
\cline{6-19} 
 &  &  &  &  & \multirow{2}{*}{Partition} & \multirow{2}{*}{	Expansion } & \multirow{2}{*}{Total} & \multirow{2}{*}{TG} & \multirow{2}{*}{LS} & \multirow{2}{*}{MA} & \multirow{2}{*}{BA} & \multirow{2}{*}{Total} & \multicolumn{3}{c|}{Original} & \multicolumn{3}{c}{Ours}\tabularnewline
\cline{14-19} 
 &  &  &  &  &  &  &  &  &  &  &  &  & TG & MA & BA & TG & MA & BA\tabularnewline
\hline 
City A & 1210106 & 50 Mpixel & 164.8k & 23867 & 25.24 & 18.84 & 46.88 & 59.02 & 34.62 & 75.26 & 56.04 & 275.74 & 2933.76 & 39.81 & 10159.62 & 34.62 & 39.81 & 0.53\tabularnewline
\hline 
City B & 138200 & 24 Mpixel & 73.0k & 2721 & 6.62 & 4.61 & 11.71 & 5.73 & 3.62 & 7.34 & 6.24 & 23.43 & 207.76 & 4.59 & 666.92 & 16.47 & 4.59 & 0.63\tabularnewline
\hline 
City C & 91732 & 50 Mpixel & 170.1k & 1723 & 5.12 & 3.17 & 8.62 & 2.64 & 2.30 & 4.27 & 7.76 & 18.10 & 162.50 & 3.04 & 492.39 & 12.33 & 3.04 & 0.62\tabularnewline
\hline 
City D & 36480 & 36 Mpixel & 96.4k & 635 & 2.01 & 1.25 & 3.57 & 1.11 & 1.21 & 1.71 & 3.31 & 7.64 & 55.70 & 1.21 & 176.57 & 4.87 & 1.21 & 0.67\tabularnewline
\hline 
\end{tabular}
}
\end{center}
\vspace{-4mm}
\caption
{
Statistics of the city-scale data-sets.
$\bar{N}_{f}$ and $N_{k}$ denote the average number of features per image and the number of camera clusters.
We abbreviate track generation, local SfM, motion averaging and full bundle adjustment to TG, LS, MA and BA respectively.
The original peak memory is an estimation of the different steps of the standard SfM pipeline~\cite{moulon2013} if handled by a single computer.
}
\label{tab:cityscale}
\vspace{-6mm}
\end{table*}

\begin{table}[t!]
\begin{center}
\resizebox{1.0\linewidth}{!}
{
\begin{tabular}{c|>{\centering}p{1.6cm}|>{\centering}p{1.8cm}|>{\centering}p{1.6cm}|>{\centering}p{1.6cm}|>{\centering}p{2.0cm}}
\hline 
Data-set & Resolution {[}Mpixels{]} & \# registered cameras  & \# tracks & Avg. track length & Avg. reproj. error {[}pixels{]}\tabularnewline
\hline 
Theia~\cite{theia-manual} & 2.25 & 19,014 & 6.78M & 4.6 & 1.84\tabularnewline
\hline 
OpenMVG~\cite{moulon2013} & 2.25 & 13,254 & 4.21M & 4.9 & 1.67\tabularnewline
\hline 
VisualSfM~\cite{wu2013} & 2.25 & 7,230 & 2.64M & 4.3 & 0.88\tabularnewline
\hline 
Colmap~\cite{schoenberger2016sfm} & 2.25 & 21,431 & 5.75M & 5.2 & 0.86\tabularnewline
\hline 
Ours & 36.15 & 36,428 & 27.8M & 6.2 & 1.18\tabularnewline
\hline 
\end{tabular}
}
\end{center}
\vspace{-2mm}
\caption
{
Comparison with the standard SfM pipelines~\cite{moulon2013,schoenberger2016sfm,theia-manual,wu2013} on City-A data-set (36480 images).
We resize images from 36 megapixels to 2.25 megapixels to fit City-D data-set into the standard pipeline~\cite{moulon2013,schoenberger2016sfm,theia-manual,wu2013}. Our approach uses the images of the original resolution.
}
\label{tab:cityscale_downsample}
\vspace{-2mm}
\end{table}

\paragraph{City-scale data-sets}\vspace{-5mm}
The statistics of the input city-scale data-sets are shown in Table~\ref{tab:cityscale}.
The image resolution ranges from $24$ to $50$ megapixels and the average number of detected features of each image ranges from 73.0K to 170.1K.
We can see that the estimated peak memory of the largest City-A data-set is 2.9TB, 39.81GB, and 10.2TB in track generation, motion averaging and bundle adjustment respectively if handle by the standard SfM pipeline~\cite{moulon2013} in a single computer,
which obviously runs out of memory of our servers with 128GB memory.
The same goes for the other standard SfM pipelines~\cite{schoenberger2016sfm,theia-manual,wu2013}.
However, our pipeline can even recover 1.21 million accurate and consistent camera poses and 1.68 billion sparse 3D points of the largest City-A data-set.
The corresponding peak memory dramatically drops to 34.62GB and 0.53GB in track generation and bundle adjustment respectively.
In Figure~\ref{fig:cityscale}, we further provide the visual results of the city-scale data-sets containing both mesh and textured models with delicate details to qualitatively demonstrate the high accuracy of the finally recovered camera poses.
As shown in Table~\ref{tab:cityscale_downsample}, we fit the whole City-D data-set to the standard SfM pipeline~\cite{moulon2013,schoenberger2016sfm,theia-manual,wu2013} by resizing images.
We can see that down-sampling images leads to an obviously smaller number of registered cameras.

\paragraph{Running time}\vspace{-5mm}
We test the Internet data-set~\cite{wilson2014} on a single computer to make a fair comparison on running time,
and Table~\ref{tab:internet} shows that our efficiency is comparable to the works in~\cite{jiang2013,ozyesil2015,theia-manual,wilson2014}.
As for the city-scale data-sets, we note in Table~\ref{tab:cityscale} that the running time of track generation and local incremental SfM grows linearly as the number of images increases,
while the running time of bundle adjustment,
the complexity of which is $\mathcal{O}((m+n)^3)$ given $m$ cameras and $n$ 3D points even in a distributed manner,
and motion averaging that can only be handled in a single computer gradually dominates as the number of images drastically increases.
Even for the City-B data-set, 
our parallel computing system composed of 10 computers can successfully reconstruct 138 thousand cameras and 100 million sparse 3D points within one day.
Notably, because of the concise design of our clustering algorithm,
the range of its running time on the city-scale data-sets is from 3.57 to 11.71 minutes,
which is extremely efficient compared with the time cost of the whole SfM pipeline.

\paragraph{Limitations}\vspace{-4mm}
Thanks to the fully scalable formulation of our SfM pipeline in terms of camera clusters,
the peak memory of track generation of our pipeline is only 2.1\%-8.7\% of the standard pipeline~\cite{cui2015,moulon2013,snavely2006,theia-manual,wu2013},
and the peak memory of bundle adjustment of our approach is even 0.1-3.8\textperthousand\;of the standard pipeline.
However, since our motion averaging formulation (Section~\ref{sec:motion_averaging}) still solves all the camera poses considering available relative motions at once,
it is limited by the memory of a single computer.
We are therefore interested to exploit our scalable formulation to solve large-scale motion averaging problems in a scalable and parallel manner,
and leave this for future study.

\vspace{-2mm}
\section{Conclusions}
\vspace{-1mm}
In this paper, we propose a parallel pipeline able to handle accurate and consistent SfM problems far exceeding the memory of a single computer.
A graph-based camera clustering algorithm is first introduced to divide the original problem into sub-problems while preserving sufficient connectivities among cameras for a highly accurate and consistent reconstruction.
A hybrid SfM method embracing the advantages of both incremental and global SfM methods is subsequently proposed to merge partial reconstructions into a globally consistent reconstruction.
Our pipeline is able to handle city-scale SfM problems containing one data-set with 1.21 million high-resolution images, 
which runs out of memory in the available approaches, 
in a highly scalable and parallel manner with superior accuracy and consistency over the state-of-the-art methods.

{\small
\bibliographystyle{ieee}
\bibliography{egbib} 
}

\end{document}